\documentclass{article}
\usepackage{ijcai20,graphicx,times,amsmath,amssymb,subfigure,threeparttable,booktabs,multirow}
\usepackage[lined,ruled,commentsnumbered]{algorithm2e}
\usepackage[hidelinks]{hyperref}
\usepackage[utf8]{inputenc}
\usepackage[small]{caption}
\allowdisplaybreaks 

\makeatletter
\newcommand{\tabincell}[2]{\begin{tabular}{@{}#1@{}}#2\end{tabular}}
\makeatother

\begin{document}

\title{Pool-Based Unsupervised Active Learning for Regression\\ Using Iterative Representativeness-Diversity Maximization (iRDM)}

\author{
Ziang Liu$^1$\and
Xue Jiang$^1$\and
Hanbin Luo$^2$\and
Weili Fang$^2$\and
Jiajing Liu$^2$\and
Dongrui Wu$^1$\footnote{Contact Author. This article has been submitted to Pattern Recognition Letters.}\\
\affiliations
$^1$Key Laboratory of the Ministry of Education for Image Processing and Intelligent Control,\\ School of Artificial Intelligence and Automation, Huazhong University of Science and Technology, China\\
$^2$School of Civil Engineering and Mechanics, Huazhong University of Science and Technology, China\\
\emails
\{ziangliu, xuejiang, luohbcem, weili\_f, liu\_jiajing, drwu\}@hust.edu.cn}

\maketitle

\begin{abstract}
Active learning (AL) selects the most beneficial unlabeled samples to label, and hence a better machine learning model can be trained from the same number of labeled samples. Most existing active learning for regression (ALR) approaches are supervised, which means the sampling process must use some label information, or an existing regression model. This paper considers completely unsupervised ALR, i.e., how to select the samples to label without knowing any true label information. We propose a novel unsupervised ALR approach, iterative representativeness-diversity maximization (iRDM), to optimally balance the representativeness and the diversity of the selected samples. Experiments on 12 datasets from various domains demonstrated its effectiveness. Our iRDM can be applied to both linear regression and kernel regression, and it even significantly outperforms supervised ALR when the number of labeled samples is small.
\end{abstract}

\section{Introduction}

In many practical regression problems, unlabeled data can be easily obtained; however, it may be very time-consuming and/or expensive to label them. For example, in emotion estimation from speech signals, it is easy to record a large number of speech utterances; however, multiple assessors are needed to evaluate the emotion primitives (e.g., 6-17 in the VAM corpus \cite{Grimm2008}, and at least 110 in IADS-2 \cite{Bradley2007}), which is labor-intensive.

Active learning (AL) is frequently used to reduce the labeling effort in such applications. Many excellent AL approaches have been proposed, however, most of them are designed for classification problems \cite{Abe1998,Cai2014,gal2017deepbayesianAL,Krogh1995,Settles2008,Settles2008b}.

Researches of active learning for regression problems (ALR) are limited.
There are two ALR scenarios: \emph{population-based} and \emph{pool-based} \cite{Sugiyama2009}. This paper considers the latter, where a pool of unlabeled samples is available, and we need to optimally select a smaller number of them to label, so that a regression model trained from them can achieve the best possible performance.

\begin{figure*}[!h]\centering
\includegraphics[width=.82\linewidth,clip]{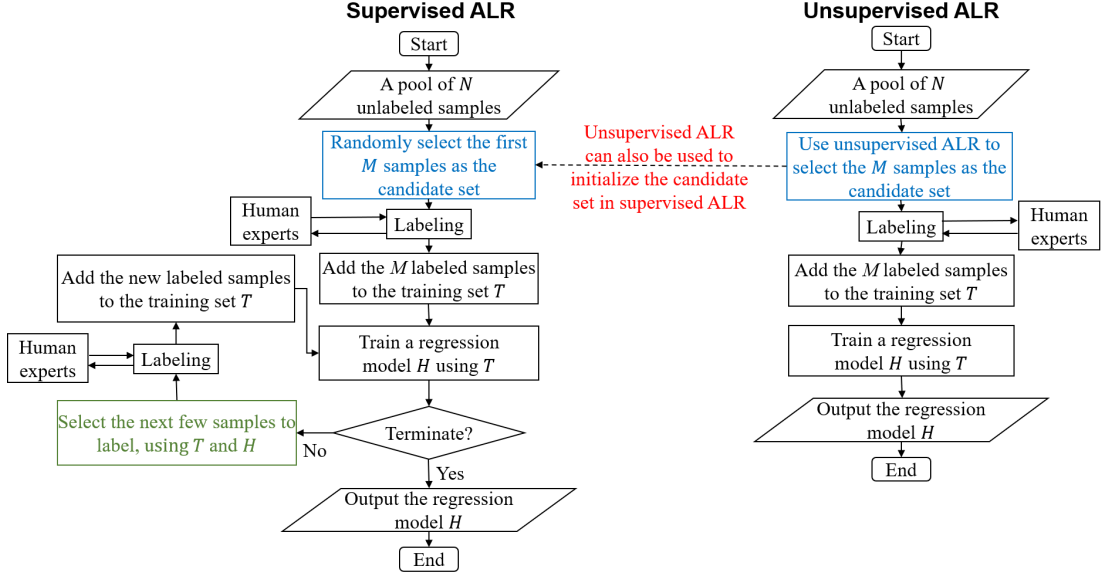}
\caption{Flowcharts of supervised ALR and unsupervised ALR.} \label{fig:ALR}
\end{figure*}

Most existing pool-based ALR approaches are supervised \cite{Burbidge2007c,Cai2013,elreedy2019entropyALR,drwuSAL2019,drwuiGS2019,Yu2010}, where some true labels are needed to guide sample selection. Only few studies explicitly considered completely unsupervised ALR, where sample selection is implemented without any label information. The differences between supervised ALR and unsupervised ALR are illustrated in Figure~\ref{fig:ALR}.

Unsupervised ALR is needed in many real-world applications. For example, in cold start of supervised ALR, where no label information is available at all, unsupervised ALR must be used to select the very first few samples to label and to build a good initial regression model.

Unsupervised ALR may also be advantageous to supervised ALR in certain applications. For example, supervised ALR needs to interact with the labeler continuously: in each iteration, supervised ALR selects some unlabeled samples, asks the labeler to label them, and then updates the regression model. This requires the labeler to be available when the ALR algorithm is running, which may be difficult sometimes. On the contrary, unsupervised ALR selects all candidate samples at once, and the labeler can work on them any time.

This paper considers pool-based unsupervised ALR. It makes the following contributions:
\begin{enumerate}
\item We propose a novel unsupervised ALR approach, iterative representativeness-diversity maximization (iRDM), which optimally balances the representativeness and the diversity of the selected samples for labeling.
\item Our proposed iRDM can be used for both linear regression and kernel regression, whereas most existing ALR approaches can only be used for linear regression.
\item We demonstrate the effectiveness of iRDM on various datasets from diverse application domains, and show that it even outperform supervised ALR when the number of labeled samples is small.
\end{enumerate}

\section{Improved Representativeness-Diversity Maximization (iRDM)} \label{sect:iRDM}

This section introduces our proposed iRDM approach for pool-based unsupervised ALR, which considers the following problem: The pool consists of $N$ unlabeled samples $\{\mathbf{x}_n\}_{n=1}^N$; we need to optimally select $M$ from it (these $M$ samples form a \emph{candidate set}), ask the labeler to label them, and then train a regression model to label future unknown samples. ``\emph{Unsupervised}" means the selection of the $M$ samples is completely unsupervised: no label information is available at all.

First, iRDM uses the RD algorithm \cite{drwuSAL2019} to select $M$ samples as the initial candidate set. More specifically, it performs $k$-means clustering ($k=M$) on all samples in the pool, and selects one sample from each cluster, which is closest to the corresponding cluster centroid. The RD strategy can achieve a good compromise between the representativeness (the samples are close to the cluster centroids, so they are unlikely to be outliers) and the diversity (the samples are from different clusters, so they cover the entire input space), but it can still be improved.

\begin{figure}[!h]\centering
\includegraphics[width=.96\linewidth,clip]{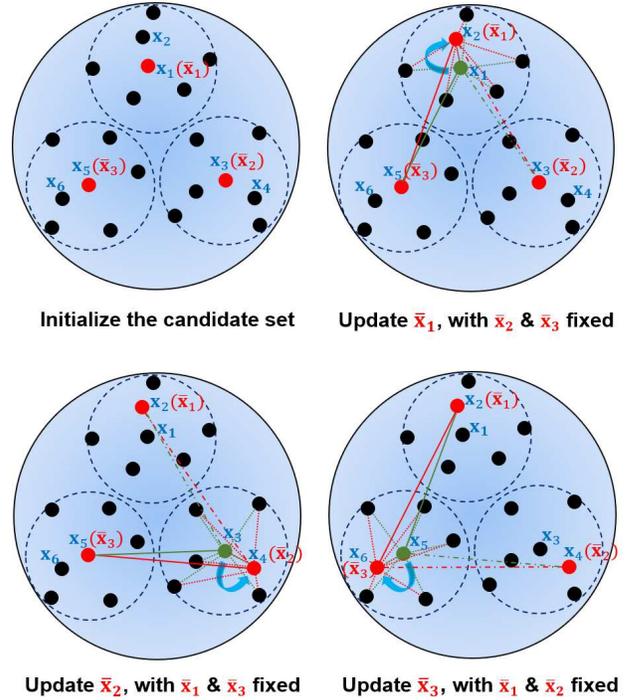}
\caption{Optimization of the candidate set in iRDM ($M=3$, one iteration). The blue dashed circles represent the cluster boundaries. In each subplot, the dashed lines represent the distances of the sample under consideration to other samples in the same cluster, whose average is $R$ in (\ref{eq:iRDM_R}). The solid and dotted lines represent the distances of the sample under consideration to the $M-1$ fixed samples in the candidate set, among which the solid line is the shortest and is $D$ in (\ref{eq:iRDM_D}). The green and red dots are the samples before and after optimization, respectively.} \label{fig:iRDM}
\end{figure}

Further optimizing the selection of the $M$ samples is an $M$-objective problem, which may not be solved easily. Inspired by the Expectation-Maximization (EM) algorithm, we break it down into $M$ single-objective optimization problems, as shown in Figure~\ref{fig:iRDM}. In each such problem, only one candidate sample is optimized while the other $M-1$ samples are fixed.

Let the current candidate set be $\{\bar{\mathbf{x}}_m\}_{m=1}^M$ (note that $\bar{\mathbf{x}}_m$ represents the $m$the sample in the candidate set, instead of the $m$th sample in the pool). Let the candidate sample to be optimized be $\bar{\mathbf{x}}_m$, and its corresponding cluster be $C_m$. Assume there are $N_m$ samples in $C_m$. iRDM selects a better sample in $C_m$ to replace $\bar{\mathbf{x}}_m$.

To measure the representativeness, for each sample $\mathbf{x}_n$ in $C_m$, we calculate the average distance between $\mathbf{x}_n$ and all remaining samples in $C_m$:
\begin{align}
R(\mathbf{x}_n) = \frac{1}{N_m-1}\times\sum_{\mathbf{x}_i\in C_m}||\mathbf{x}_n-\mathbf{x}_i|| \label{eq:iRDM_R}
\end{align}

To measure the diversity, inspired by GSx \cite{drwuiGS2019}, for each sample $\mathbf{x}_n$ in $C_m$, we calculate the distance between $\mathbf{x}_n$ and each of the $M-1$ fixed samples in the candidate set, and use the minimum distance to measure the diversity of $\mathbf{x}_n$:
\begin{align}
D(\mathbf{x}_n)=\min_{i\in[1,M]\atop i\neq m}||\mathbf{x}_n-\mathbf{x}_i|| \label{eq:iRDM_D}
\end{align}

We calculate $R(\mathbf{x}_n)$ and $D(\mathbf{x}_n)$ for each $\mathbf{x}_n$ in $C_m$, and select the optimal sample $\mathbf{x}^*_n$ with the maximum objective value to replace $\bar{\mathbf{x}}_m$ in the candidate set:
\begin{align}
\mathbf{x}^*_n=\arg\max_{n\in[1,N_C]} [D(\mathbf{x}_n)-R(\mathbf{x}_n)] \label{eq:iRDM_RD}
\end{align}

This completes the single-task optimization for $\bar{\mathbf{x}}_m$. We then move on to update the next sample in the candidate set. One iteration is done if all $M$ samples in the candidate set have been updated once. iRDM terminates when the samples in the candidate set stop change, or the maximum number of iterations is reached, as shown in Algorithm~\ref{alg:1}.

\begin{algorithm}[h] 
\KwIn{A pool of $N$ unlabeled samples, $\{\mathbf{x}_n\}^N_{n=1}$\;
    \hspace*{10mm} $c_{\max}$, the maximum number of iterations.}
\KwOut{$\{\bar{\mathbf{x}}_m\}_{m=1}^M$, the set of $M$ samples to label.}
Perform $k$-means clustering ($k=M$) on $\{\mathbf{x}_n\}^N_{n=1}$, and denote the clusters as $\{C_m\}_{m=1}^M$\;
Select $\mathbf{x}_m$ as the sample closest to the centroid of $C_m$, $m=1,...,M$\;
Sort the indices of the $M$ samples in the candidate set and save them to the first row of matrix $P$\;
Compute $R(\mathbf{x}_n)$ in (\ref{eq:iRDM_R}) for $n=1,...,N$ and save them\;
$c=0$\;
\While{$c<c_{\max}$}{
    Denote the $M$ selected samples as $\{\bar{\mathbf{x}}_m\}_{m=1}^M$\;  
    \For{$m=1,...,M$}{
        Fix $\{\mathbf{x}_1,\ldots,\mathbf{x}_{m-1},\mathbf{x}_{m+1},\ldots,\mathbf{x}_M\}$\;
        Compute $D(\mathbf{x}_n)$ in (\ref{eq:iRDM_D}) for each sample in $C_m$\;
        Identify $\mathbf{x}^*_n$ in (\ref{eq:iRDM_RD})\;
        Set $\bar{\mathbf{x}}_m$ to $\mathbf{x}^*_n$\;}
        Sort the indices of the $M$ samples in the candidate set\;
    \uIf{the sorted indices of the $M$ samples match any row in $P$}{
        \textbf{Break}\;}
    \Else{
        Save the sorted indices of the $M$ samples to the next row of $P$\;}
            $c=c+1$\;}
\caption{The proposed iRDM algorithm.} \label{alg:1}
\end{algorithm}

\section{Experimental Results}\label{sect:experiments}

Extensive experiments are performed in this section to demonstrate the performance of the proposed iRDM.

\subsection{Datasets} \label{sect:Datasets}

A summary of the 12 datasets used in our experiments are shown in Table~\ref{tab:Datasets}. They cover a wide variety of application domains. Eleven datasets are from the UCI Machine Learning Repository\footnote{http://archive.ics.uci.edu/ml/index.php} and the CMU StatLib Datasets Archive\footnote{http://lib.stat.cmu.edu/datasets/}, which have also been used in many previous ALR experiments \cite{Cai2017,Cai2013,Yu2010,drwuSAL2019,drwuiGS2019}. We also used an affective computing dataset: \emph{Vera am Mittag} (VAM; \emph{Vera at Noon} in English) \cite{Grimm2008}), which has been used in many previous studies \cite{Grimm2007b,Grimm2007a,drwuMTALR2018}. Only arousal in VAM was used as the regression output.

\begin{table}[!h] \centering  \setlength{\tabcolsep}{1mm} \footnotesize
\caption{Summary of the 12 regression datasets.}   \label{tab:Datasets}
\begin{tabular}{c|cccccc}   \hline
Dataset     &\tabincell{c}{No. of\\samples} &\tabincell{c}{No. of\\raw\\features} &\tabincell{c}{No. of\\numerical\\features} &\tabincell{c}{No. of\\categorical\\features} &\tabincell{c}{No. of\\total\\features}    \\ \hline
Concrete-CS    &103            &7                   &7                         &0           &7\\
Yacht          &308            &6                   &6                         &0           &6\\
autoMPG        &392            &7                   &6                         &1           &9\\
NO2            &500            &7                   &7                         &0           &7\\
Housing        &506            &13                  &13                        &0          &13 \\
CPS            &534            &10                  &7                         &3          &19\\
EE-Cooling     &768            &7                   &7                         &0          &7\\
VAM-Arousal     &947            &46                  &46                        &0         &46\\
Concrete        &1,030           &8                   &8                         &0         &8\\
Airfoil       &1,503           &5                   &5                         &0        &5\\
Wine-Red      &1,599           &11                  &11                        &0        &11\\
Wine-White    &4,898           &11                  &11                        &0         &11\\ \hline
\end{tabular}
\end{table}

Two datasets (autoMPG and CPS) contain both numerical and categorical features. For them, we used one-hot encoding to covert the categorical values into numerical values before ALR. For each dataset, we normalized each dimension of the input to mean zero and standard deviation one.

\subsection{Algorithms}

We compared iRDM ($c_{\max}=5$) with the following nine sampling approaches, including four unsupervised sampling approaches:
\begin{enumerate}
\item Random sampling (RS), which randomly selects $M$ samples for labeling.
\item P-ALICE (Pool-based Active Learning using the Importance-weighted least-squares learning based on Conditional Expectation of the generalization error) \cite{Sugiyama2009}, which estimates the label uncertainty as the weights while selecting the $M$ samples, and builds a weighted linear regression model from them. The parameter $\lambda$ was chosen as the best one from $\{0,.1,.2,.3,.4,.41,.42,...,.59,.6,.7,.8,.9,1\}$, as in \cite{Sugiyama2009}.
\item GSx (Greedy Sampling in the Input Space) \cite{Yu2010,drwuiGS2019}, which maximizes the diversity of the $M$ selected samples in the feature space.
\item RD (Representativeness-Diversity) \cite{drwuSAL2019}, which performs $k$-means clustering ($k=M$) on the $N$ unlabeled samples, and selects from each cluster one sample closest to the cluster centroid. RD considers sample representativeness and diversity simultaneously.
\end{enumerate}
and five supervised sampling approaches:
\begin{enumerate}
\item QBC (Query-By-Committee) \cite{RayChaudhuri1995}, which selects the sample with the maximum variance computed from a committee of regression models. Four regression models were used, and random sampling was used to select the first five samples.
\item EMCM (Expected Model Change Maximization) \cite{Cai2013}, which selects the sample that will change the model parameters the most. Four regression models were used, and random sampling was used to select the first five samples. It only applies to linear regression.
\item RD-EMCM \cite{drwuSAL2019}, which integrates RD and EMCM. RD was used to select the first five samples.
\item iGS (improved Greedy Sampling) \cite{drwuiGS2019}, which uses greedy sampling in both the feature space and the label space. GSx was used to select the first sample.
\item RSAL (Residual regression) \cite{douak2013KRRAL}, which predicts the estimation error of each unlabeled sample and selects the one with the largest error to label. Random sampling was used to select the first five samples.
\end{enumerate}

Note that RS, GSx, RD, QBC and iRDM apply to both linear and kernel regression. P-ALICE, EMCM, RD-EMCM and iGS only apply to linear regression. RSAL only applies to kernel regression.

\subsection{Performance Evaluation Process}\label{sect:evaluation}

For each dataset, we randomly select $50\%$ samples as the training pool, and the remaining $50\%$ as the test set. We used the mean and the variance of the training samples to normalize the test samples, because in practice the test samples are unknown.

Each sampling approach selected $M\in[5, 50]$ samples from the training pool to label, and then built a regression model, which was evaluated on the test set. The performance measures were the root mean squared error (RMSE) and the correlation coefficient (CC). This process was repeated 100 times to obtain statistically meaningful results.

The following two regression models were used:
\begin{enumerate}
\item Ridge regression (RR), with the regularization coefficient $r=0.1$.
\item Radial basis function support vector regression (RBF-SVR), with the box constraint for the alpha coefficients $C=50$, and the half width of the epsilon-insensitive band $\epsilon=0.1\sigma(Y)$, where $\sigma(Y)$ is the standard deviation of the true labels of the $M$ selected samples.
\end{enumerate}
Note that RR was used for RS, GSx, iGS, QBC, EMCM, RD, RD-EMCM and iRDM, and RBF-SVR was used for RS, GSx, RD, QBC, RSAL and iRDM.

\subsection{Experimental Results} \label{sect:}

\begin{figure*}[!h]\centering
\includegraphics[width=\linewidth,clip]{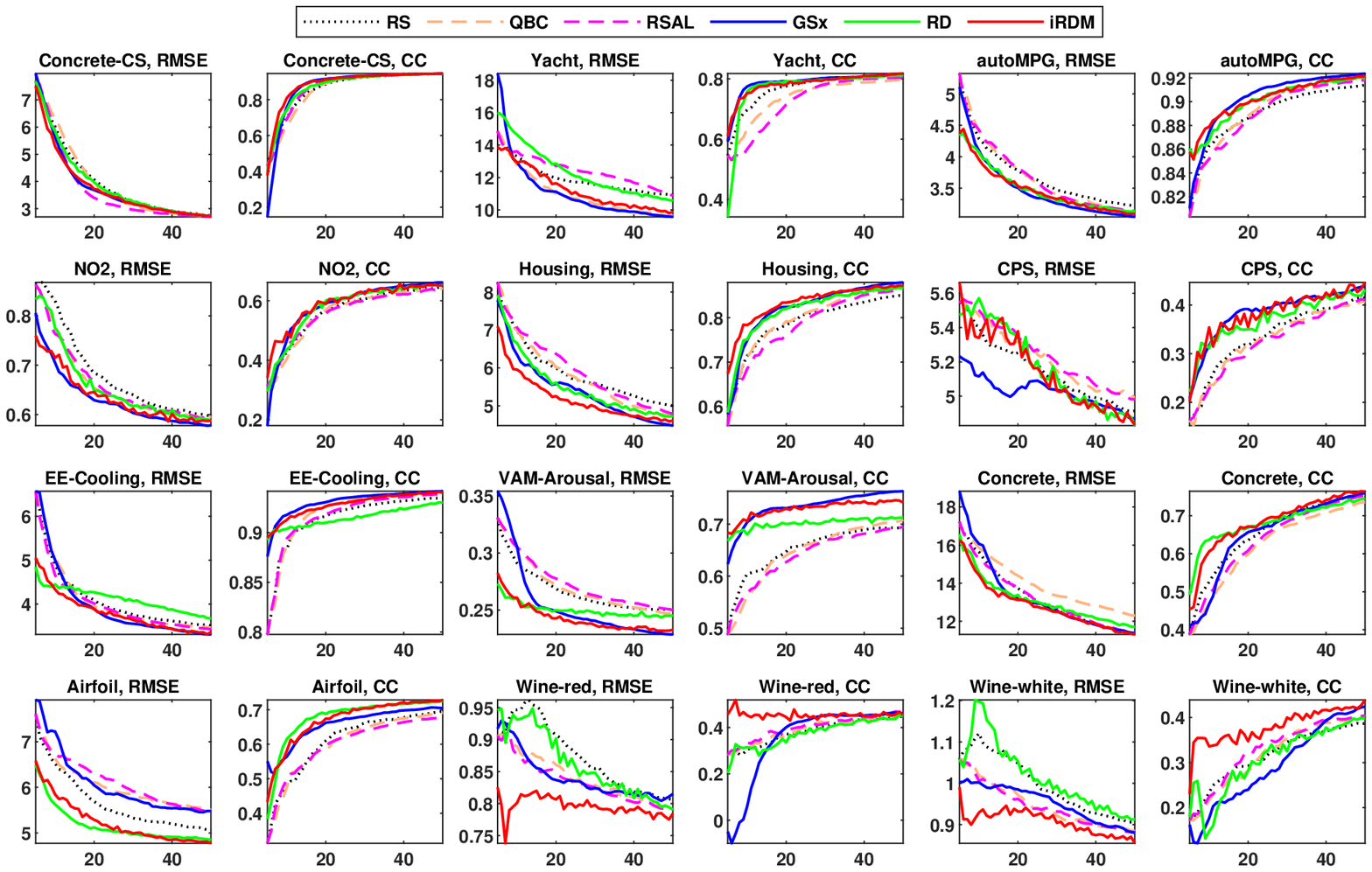}
\caption{Mean of the RMSEs and the CCs on the 12 datasets, averaged over 100 runs. The horizontal axis represents $M$, the number of samples to be labeled. RBF-SVR ($C=50$, $r=0.01$) was used as the regression model.} \label{fig:curve_RBFSVR}
\end{figure*}

Due to the page limit, we only show the detailed results from RBF-SVR in Figure~\ref{fig:curve_RBFSVR}. iRDM performed the best on most datasets and for most $M$.

\begin{figure}[!h]\centering
\includegraphics[width=\linewidth,clip]{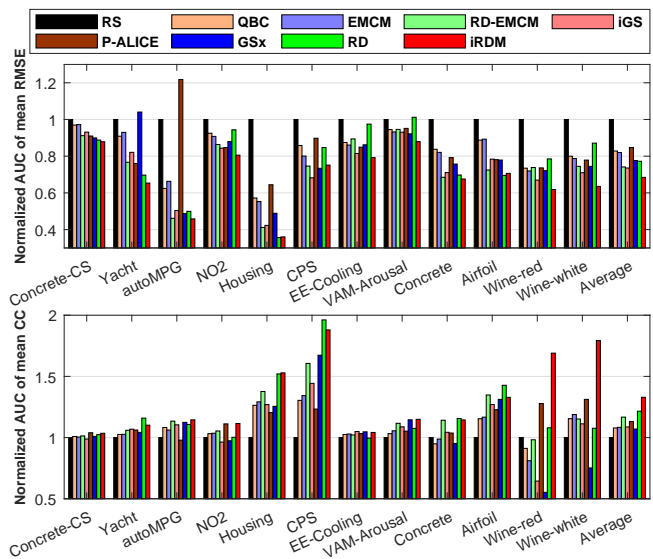}
\caption{Normalized AUCs ($M\in[5,20]$) of the mean RMSEs and the mean CCs on the 12 datasets. RR ($r=0.1$) was used.} \label{fig:RR2}
\end{figure}

\begin{figure}[!h]\centering
\includegraphics[width=\linewidth,clip]{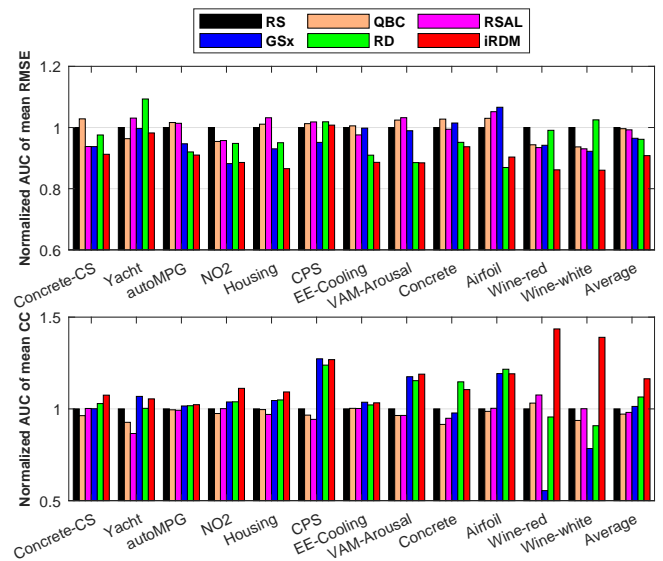}
\caption{Normalized AUCs ($M\in[5,20]$) of the mean RMSEs and the mean CCs on the 12 datasets. RBF-SVR ($C=50$, $\lambda=0.01$) was used.} \label{fig:RBFSVR}
\end{figure}

To see the forest for the trees, we also computed the area under the curves (AUCs) of the mean RMSEs and the mean CCs for different regression models. Since the benefits of active learning (in both classification and regression) vanish when $M$ becomes large (thus active learning is usually used for small $M$), we used $M\in[5,20]$ to compute the AUC value for each approach. The results for RR and RBF-SVR are shown in Figures~\ref{fig:RR2} and \ref{fig:RBFSVR}, respectively. Because AUCs from different datasets varied significantly, we normalized them w.r.t. the AUC of RS on each dataset, thus the AUC of RS on each dataset was always 1. On average, iRDM performed the best across the 12 datasets, for both RMSE and CC, and for both RR and RBF-SVR.

To quantify the performance improvements of ALR approaches over RS, we computed the percentage improvements on the AUCs of the RMSEs and the CCs, as shown in Table~\ref{tab:iRDM_8AlgsvsRS_linear} for RR and Table~\ref{tab:iRDM_5AlgsvsRS_kernel} for RBF-SVR. On average, iRDM had the largest performance improvements, for both RMSE and CC.

\begin{table}[h] \setlength{\tabcolsep}{0.5mm} \footnotesize
  \centering
  \caption{Percentage improvements of the AUCs of the mean RMSEs and the mean CCs over RS. RR was the regression model. The best performances are marked in bold.}
    \begin{tabular}{cc|cccccccc}
    \toprule
      &   & QBC & EMCM & \multicolumn{1}{c}{\tabincell{c}{RD-\\EMCM}} & iGS & \multicolumn{1}{c}{\tabincell{c}{P-\\ALICE}} & GSx & RD & iRDM \\
    \midrule
    \multirow{2}[2]{*}{RMSE} & Mean & 17.2  & 18.0  & 25.9  & 26.5  & 15.3  & 22.4  & 22.8  & \textbf{31.5}  \\
      & Var & 49.9  & 50.9  & 76.3  & 81.5  & 57.6  & 81.3  & 72.0  & \textbf{87.5}  \\
    \midrule
    \multirow{2}[2]{*}{CC} & Mean & 7.9  & 8.3  & 16.7  & 8.7  & 13.1  & 7.0  & 21.5  & \textbf{32.9}  \\
      & Var & 13.2  & 17.7  & 46.5  & 33.2  & 31.9  & 35.0  & 56.1  & \textbf{64.1}  \\
    \bottomrule
    \end{tabular}%
  \label{tab:iRDM_8AlgsvsRS_linear}%
\end{table}%

\begin{table}[!h]
  \centering
  \caption{Percentage improvements of the AUCs of the mean RMSEs and the mean CCs over RS. RBF-SVR was the regression model. The best performances are marked in bold.}
    \begin{tabular}{cc|ccccc}
    \toprule
      &   & QBC & RSAL & GSx & RD & iRDM \\
    \midrule
    \multirow{2}[2]{*}{RMSE} & Mean & 0.4  & 0.8  & 3.5  & 3.9  & \textbf{9.2}  \\
      & Var & -5.7  & -1.1  & -7.4  & 46.8  & \textbf{51.0}  \\
    \midrule
    \multirow{2}[2]{*}{CC} & Mean & -2.8  & -1.9  & 1.3  & 6.5  & \textbf{16.4}  \\
      & Var & -16.1  & -7.9  & 47.9  & 46.2  & \textbf{58.9}  \\
    \bottomrule
    \end{tabular}%
  \label{tab:iRDM_5AlgsvsRS_kernel}%
\end{table}%

Interestingly, when the number of labeled samples was small ($M\in[5, 20]$), iRDM even outperformed supervised ALR approaches (QBC, EMCM, RD-EMCM, iGS, and RSAL), since the latter may be misled by inaccurate regression models trained from very few labeled samples.

\subsection{Statistical Analysis} \label{sect:stat_analysis}

To determine if the differences between iRDM and other ALR approaches were statistically significant, we also performed non-parametric multiple comparison tests on them using Dunn's procedure \cite{Dunn1961}, with a $p$-value correction using the False Discovery Rate method \cite{Benjamini1995}. The results for iRDM versus the unsupervised and supervised ALR approaches are shown in Table~\ref{tab:iRDM stat_UALR_SALR}, where the statistically significant ones are marked in bold. The performance improvements of iRDM over other approaches were almost always statistically significant or very close to the boundary, regardless of the regression model and the performance measure.

\begin{table}[htpb]   \centering
  \caption{$p$-values of non-parametric multiple comparisons on the AUCs of the RMSEs and the CCs ($\alpha=0.05$; reject $H_0$ if $p<\alpha/2$) of iRDM versus other \emph{unsupervised} and \emph{supervised} sampling approaches. The statistically significant ones are marked in bold.}
    \begin{tabular}{l|cc|cc}     \toprule
    & \multicolumn{2}{c|}{RR} &\multicolumn{2}{c}{RBF SVR} \\ \cline{2-5}
    & RMSE & CC & RMSE & CC\\ \midrule
    RS & \textbf{.0000} & \textbf{.0000} & \textbf{.0002} & \textbf{.0000}\\
    QBC  & \textbf{.0000}  & \textbf{.0000} & \textbf{.0001} & \textbf{.0000}\\
    EMCM  & \textbf{.0000} & \textbf{.0000} & -- & --\\
    RD-EMCM & \textbf{.0135}  & \textbf{.0005} & -- & --\\
    iGS & \textbf{.0133} & \textbf{.0000} & -- & --\\
    RSAL & -- & --  & \textbf{.0004} & \textbf{.0000}\\
    P-ALICE & \textbf{.0000} & \textbf{.0000} &-- &--\\
    GSx & \textbf{.0002} & \textbf{.0000} & \textbf{.0191} & \textbf{.0000}\\
    RD & \textbf{.0005} & \textbf{.0198} & .0280 & \textbf{.0007}\\
    \bottomrule
    \end{tabular}%
  \label{tab:iRDM stat_UALR_SALR}%
\end{table}%

\subsection{Hyper-Parameter Sensitivity}

iRDM in Algorithm~1 has a hyper-parameter $c_{\max}$, the maximum number of iterations. This subsection studies how sensitive iRDM is to $c_{\max}$.

\begin{figure}[!h] \centering
\subfigure[]{  \includegraphics[width=.9\linewidth,clip]{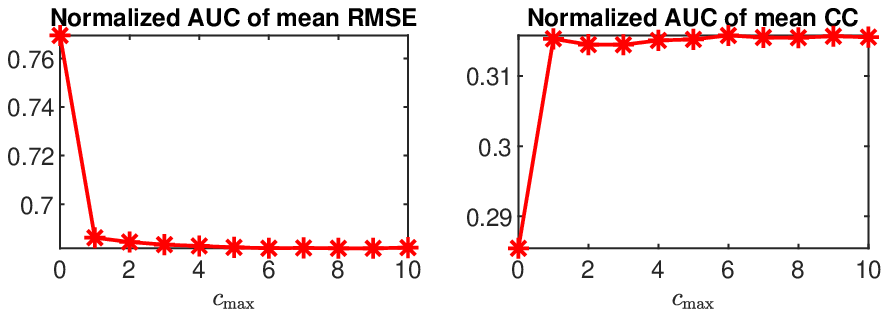}}
\subfigure[]{  \includegraphics[width=.9\linewidth,clip]{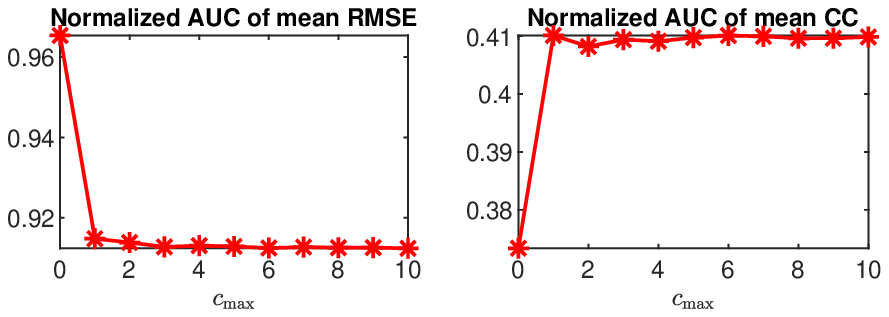}}
\caption{Ratios of AUCs of the mean RMSEs w.r.t those of RS, for iRDM with different $c_{\max}$, averaged across the 12 datasets and 100 runs. (a) RR was the regression model; (b) RBF-SVR was the regression model.} \label{fig:curve_cmax}
\end{figure}

Figure~\ref{fig:curve_cmax} shows the normalized AUCs of iRDM w.r.t. that of RS, averaged across 100 runs and 12 datasets, using RR and RBF-SVR. Note that iRDM is equivalent to RD when $c_{\max}=0$. The performance of iRDM improved quickly as $c_{\max}$ increased and converged before $c_{\max}=5$. Interestingly, the performance of iRDM was already pretty good when $c_{\max}=1$, i.e., after only one iteration. Thus, to reduce the computational cost, it is safe to choose $c_{\max}=1$.

\subsection{Initialize Supervised ALR by Unsupervised ALR}

\begin{figure}[!h]\centering
\includegraphics[width=\linewidth,clip]{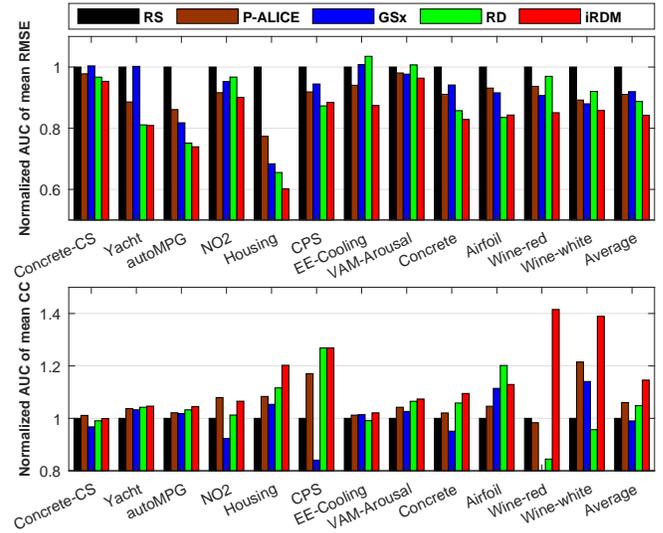}
\caption{Ratios of AUCs of the mean RMSEs w.r.t those of RS, for RD-EMCM using different unsupervised ALR to selected the first 5 samples, averaged across the 12 datasets and 100 runs.} \label{fig:bar_UALR_RD-EMCM}
\end{figure}

As shown in Figure~\ref{fig:ALR}, an unsupervised ALR can also be used to replace the random initialization in a supervised ALR to improve its performance. To verify this, we used RS, P-ALICE, GSx, RD and iRDM to initialize RD-EMCM (RR was the regression model), and show the normalized AUCs ($M\in[5,20]$) in Figure~\ref{fig:bar_UALR_RD-EMCM}. Clearly, iRDM initialization achieved the best overall performance.

\section{Conclusion}

ALR is a machine learning approach for reducing the labeling effort in regression problems. This paper considers pool-based unsupervised ALR, where the samples in a given pool are all unlabeled, and we need to select some to label without any true label information. We proposed a novel iRDM approach, which optimally balances the representativeness and the diversity of the selected samples. Experiments on datasets from various application domains demonstrated the effectiveness of iRDM. It outperformed all state-of-the-art unsupervised ALR approaches. Remarkably, it even performed better than supervised ALR approaches when the number of selected training samples is small.

The proposed iRDM can be applied to both linear regression and kernel regression. It can be used alone, or be used to better initialize the first few samples in supervised ALR.

%

\newpage

\end{document}